\documentclass{article}

     \usepackage[preprint]{neurips_2023}

\usepackage[utf8]{inputenc} 
\usepackage[T1]{fontenc}    
\usepackage{hyperref}       
\usepackage{url}            
\usepackage{booktabs}       
\usepackage{amsfonts}       
\usepackage{nicefrac}       
\usepackage{microtype}      
\usepackage{xcolor}         

\usepackage{amsfonts,amssymb}
\usepackage{amsmath}
\usepackage{amsthm}
\usepackage{float}
\usepackage{graphicx}
\usepackage{sidecap}
\usepackage{multirow}
\usepackage{caption}
\usepackage{subcaption}
\usepackage{ulem}
\usepackage{bm}
\usepackage{algorithm}
\usepackage{algorithmic}
\usepackage{natbib}
\usepackage{comment}
\usepackage{ulem}

\theoremstyle{plain}

\theoremstyle{definition}

\title{Advancing Ad Auction Realism:\\ Practical Insights \& Modeling Implications}

\author{
Ming Chen \thanks{work done while at Amazon Ads; can be reached at  \texttt{mingo.chen198906@gmail.com}}\\
Outreach Inc. \\
Seattle, USA \\
\And
Sareh Nabi \thanks{can be reached at \texttt{sareh@amazon.com}}\\
Amazon Ads \\
Seattle, USA \\
\And
Marciano Siniscalchi \thanks{can be reached at \texttt{siniscam@amazon.com} or \texttt{marciano@northwestern.edu}}\\
Amazon Ads \& Northwestern Univ. \\
Evanston, USA\\
}

\begin{document}

\maketitle

\begin{abstract}
Contemporary real-world online ad auctions differ from canonical models \citep{edelman2007internet,varian2009online} in at least four ways: (1) values and click-through rates can depend upon users' search queries, but advertisers can only partially ``tune'' their bids to specific queries; (2) advertisers do not know the number, identity, and precise value distribution of competing bidders; (3) advertisers only receive partial, aggregated feedback, and (4) payment rules are only partially known to bidders. These features make it virtually impossible to fully characterize equilibrium bidding behavior.  This paper shows that, nevertheless, one can still gain useful insight into modern ad auctions by modeling advertisers as agents governed by an adversarial bandit algorithm, independent of auction mechanism intricacies. To demonstrate our approach, we first simulate  ``soft-floor'' auctions \citep{zeithammer2019soft}, a complex, real-world pricing rule for which no complete equilibrium characterization is known. We find that (i) when values and click-through rates are query-dependent, soft floors can improve revenues relative to standard auction formats even if bidder types are drawn from the same distribution; and (ii) with distributional asymmetries that reflect relevant real-world scenario, we find that soft floors yield lower revenues than suitably chosen reserve prices, even restricting attention to a single query. We then demonstrate how to infer advertiser value distributions from observed bids for a variety of pricing rules, and illustrate our approach with aggregate data from an e-commerce website.
\end{abstract}

\raggedbottom
\section{Introduction}
\label{sec:intro}

Online ad auctions are an integral part of contemporary e-commerce. As end users interact with services or online stores such as Google, Amazon, or Facebook, advertisers compete to secure slots, aiming to display their ads to those prospective customers who are most likely to engage with them, and ultimately leading to conversions.

The prevalence of such auctions has led to the development of a large literature in both Economics and Computer Science. The vast majority of these contributions focus on relatively simple auction formats, in which one or more ad slots are sold using so-called generalized first- or second-price auctions \citep{edelman2007internet,varian2009online}. These mechanism adapt the first- and second-price mechanisms from ``textbook'' auction theory \citep{myerson1981optimal,krishna2009auction} to account for the fact that online advertising services often charge prices per click rather than per impression.\footnote{That is, the winner is charged if / when the user clicks on the winning ad, rather than when the winning ad is displayed. The key issue is that the ``click-through rates'' (probabilities that the user clicks on displayed ads) typically vary across advertisers; auction mechanisms must then be adjusted to provide incentives for advertisers to bid according to their true value for showing an ad.}

However, real-world online ad auction are highly complex.\footnote{For standard auctions, equilibrium behavior can be characterized, and conditions leading to revenue equivalence or revenue ranking can be established \citep{myerson1981optimal,milgrom1982theory}.} This often makes it  impossible to analyze equilibrium bidding behavior and compare the performance (e.g., revenues) of different auction formats. As a result, businesses selling online ad space lack principled guidance as to which auction format to adopt. 

We propose that simulating advertisers' behavior using well-understood online-learning algorithms can provide such guidance. To demonstrate this, we develop an approach that allows for the following key departure from standard auction formats and their generalized variants: 
\begin{enumerate}
    \item The value of an impression or click for a given ad (and slot), as well as its click-through rate, may vary depending on the shopper's query, identity, and history. Advertisers can choose ``targeting clauses'' to only bid on certain sets of queries, but cannot condition their bid on all possible variants of users' individual queries.
    \item The number and identity of advertisers changes from one auction to the next. As a result, not only do  advertisers not know the valuation of other bidders for the ad slots on offer; they also do not know the distribution those valuations are drawn from. 
    \item Advertisers receive only incomplete feedback about the outcome of each individual auction. In particular, they do not observe other advertisers' bids, and they only learn the market-clearing price if they win the auction.
    \item Pricing mechanisms are often incompletely specified. For instance, ad services may indicate that reserve prices are used, but not what they are, or how they are computed.\footnote{See for instance the \href{https://support.google.com/google-ads/answer/7634668?hl=en&ref_topic=24937}{documentation for Ad Thresholds} in Google Ads.}
\end{enumerate} 
To deal with the scant information bidders possess when they participate in real-world ad auctions, we choose learning algorithms that only require  feedback about \emph{individual} rewards, and are \emph{not tailored} to any specific auction format. In addition, we want our model to be able to handle auctions in which the number of participants and bid space is realistic. 
We focus on two algorithms: \texttt{Hedge} \citep{auer1995gambling} and  \texttt{EXP3-IX} \citep{kocak2014EXP3IX, lattimore2020bandit}.
These algorithms have known regret bounds, have been previously applied in the auction learning literature (albeit in simpler settings), and exemplify full- and partial-information algorithms respectively.

 In section \S \ref{sec:results:textbook}, we verify that, in symmetric, single-query environments, our approach reassuringly approximates theoretical predictions, including revenue equivalence  \citep{myerson1981optimal}. 

We then present our two  main contribution. The first is to demonstrate that simulation analysis can yield insight into auction environments that are too complex for explicit equilibrium analysis.  Specifically:
\begin{itemize}
    \item In \S\ref{sec:results:SFRP}, we illustrate that, with multi-query targeting, \emph{soft floors} \citep{zeithammer2019soft} \emph{can} lift revenues relative to textbook auction formats, even if bidders are ex-ante symmetric. (In a symmetric auction with a single query, we verify that revenue equivalence holds).
    \item \S\ref{sfrp_vs_rp} considers asymmetric, single-query auctions. We complement the results of \citet{zeithammer2019soft} by considering specifications of bidder value distributions that are not amenable to closed-form equilibrium solutions. We show that, in a variety of cases of real-world relevance, soft floors yield \emph{lower} revenues than second-price auctions with a suitably chosen (not necessarily optimal) reserve price. 
\end{itemize}
Taken together, these findings lead to a novel policy implication: soft floors are  not likely to be an effective tool to improve revenues in simple, single-query environments, but may be useful in multi-query environments.

The second contribution is to demonstrate how to use our approach to solve the inverse problem, namely inferring bidders' valuations through an iterative parameter search matching bid data\footnote{This is similar to structural estimation in empirical industrial organization. The key difference is that the structural equations are not derived from Bayesian Nash equilibrium behavior, but from the predictions of our learning model.}. 
\begin{itemize}
\item \S\ref{evals} validates our approach: for a range of artificial value distributions, we first simulate bids, and then apply the noted iterative search procedure to back out, or elicit, the underlying value distribution. We can thus compare the originally postulated distribution with the elicited one. 
\item \S\ref{prod} employs aggregate bid data from an actual production environment (a major e-commerce website) and infer bidder value distributions in both low and high traffic shopper query scenarios.
\end{itemize}
We view these results as a preliminary step towards a simulation-based empirical analysis of complex ad auction environments. Implementation code is available at this repository: \url{https://github.com/amzn/advancing-ad-auctions}.

\section{Related Literature}
\label{sec:related}

There is a vast literature in Economics that studies games from the perspective of learning agents; \citet{fudenberg1998theory} is an authoritative account (see also \citealp{fudenberg2009learning}).

In computer science, the literature on online learning in auctions can be roughly divided into two branches. The first takes the perspective of the seller and studies the design of revenue-maximizing auctions; recent studies include \citet{roughgarden2019minimizing},  \citet{guo2021robust}, and \citet{Jeunen2023}. This branch of the literature studies sellers who employ learning algorithms, but bidders are not explicitly modelled. 

The second branch instead takes the perspective of a single bidder who uses learning algorithms to guide her bidding process. \citet{weed2016online} focus on second-price auctions for a single good, and assume that the valuation can vary either stochastically or adversarially in each auction. In a similar environment, \citet{balseiro2018contextual} and \citet{han2020optimal} study contextual learning in first-price auctions, where the context is provided by the bidder's value. For auctions in which the bidder must learn her own value (as is often the case in the settings we consider), \citet{feng2018learning} proposes an improved version of the \texttt{EXP3} algorithm that attains a tighter regret bound. There is also a considerable literature that studies optimal bidding with  budget and/or ROI constraints using reinforcement-learning: e.g., \citet{wu2018budget}, \citet{ghosh2020optimal}, and references therein, and \citet{deng2023multi}.  \citet{golrezaei2021bidding} also studies the interaction between a seller and a single, budget- and ROI-constrained buyer.

Our paper differs from the above references since it models the \emph{interaction} among bidders who adopt online learning algorithms. In this sense, it is closer in spirit to the Economics literature on learning in games. In addition, unlike the present paper, the cited references assume that the winning bidder observes her valuations and payment in each period; some of these papers leverage insights that depend on specific auction format; and none of them allow for bidders targeting different clauses. 

To the best of our knowledge, the closest papers to our own are \citet{kanmaz2020using}, \citet{elzayn2022equilibria}, \citet{banchio2022artificial}, and \citet{jeunen2022learning}. The first reports on experiments using a multi-agent reinforcement-learning model in simple sequential (English) auctions for a single object, with a restricted bid space. Our analysis focuses on simultaneous bidding in scenarios that are representative of actual online ad auctions. The second focuses on position (multi-slot) auctions and, among other results, reports on experiments using no-regret learning (specifically, the \texttt{Hedge} algorithm we also use) under standard generalized second-price and Vickerey-Groves-Clarke pricing rules. Our analysis is complementary in that we allow for different targeting clauses and more complex pricing rules such as ``soft floors''. The third studies the emergence of spontaneous collusion in standard first- and second-price auctions, under the $Q$-learning algorithm \citep{watkins1989learning}).  The fourth describes a simulation environment similar to ours that is mainly intended to help train sophisticated bidding algorithms for advertisers. We differ in that we allow for bids broadly targeting multiple queries, and focus on learning algorithms that allow us to model auctions with a large number of bidders; in addition, we demonstrate how to infer values from observed bids.

\citet{feng2021convergence} establish the convergence to equilibrium of learning algorithms in first- and second-price auctions, as well as multi-slot VCG mechanisms. Our results in \S \ref{sec:results:textbook} provide an empirical counterpart to their theoretical results, but also add nuance as to the speed of convergence of different algorithms in realistic-sized auctions. \citet{hartline2015coarseBCE} establish the convergence of no-regret learning to coarse Bayes correlated equilibrium in general games with incomplete information; we leverage their results in \S \ref{sec:coarseBCE}.

\citet{nekipelov2015econometrics} proposes techniques for estimating agents' valuations in generalized second-price auctions, which stands in contrast to our method that directly utilizes agents' learning algorithms and is independent of the specific auction format. In a different direction,
\citet{rahme2021auction} study revenue maximization in auctions as a mechanism design problem, assuming a static ``no regret'' constraint for bidders, which in general differs from the no regret condition in online learning. \citet{peysakhovich2019robust} presents a method to predict player behavior in an unseen game. Unlike our approach, they base their predictions on an $\epsilon$-Bayesian Nash equilibrium instead of no-regret learning. \citet{bichler2021learning} suggests a method to calculate $\epsilon$-Bayesian Nash equilibria in sealed-bid auctions, focusing more on approximating equilibria, as opposed to directly modeling bidder learning behavior.

Finally, we mention equilibrium analyses of bidding and ad exchanges that provide results related to our simulation findings. \citet{choi2019learning} analyze auctions in which a new entrant's click-through rate is not known to the publisher. \citet{despotakis2021first} demonstrates that, with competing ad exchanges, a multi-layered auction involving symmetric bidders can result in a scenario where first-price auctions, and soft floors in general, yield higher revenue than second-price auctions. When multiple slots are offered, \citet{rafieian2021targeting} shows that, when advertisers can target their bids to  specific placements in second-price auctions, total surplus increases, but the effect on publisher revenues is ambiguous, the key difference being that we consider a single slot with randomly arriving queries, rather than multiple slots that are often offered for sale. \citet{nabi2022bayesian} propose a hierarchical empirical Bayes method that learns empirical meta-priors from the data in Bayesian frameworks. They further apply their approach in a contextual bandit setting, demonstrating improvements in performance and convergence time.

\section{Auction Model and Learning}

This section introduces the auction model we analyze. For any finite set $\mathcal{X}$, we denote by $\Delta(\mathcal{X})$ the set of probability distributions on $\mathcal{X}$.

\subsection{Auction Model}
\label{sec:model}

We fix a single ad slot, or position, within the ad service's or publisher's web page. 
Time is discrete and indexed by $t = 1,2,\ldots,T$. In each time period $t$, a user visits the web page and submits a query, represented by a point $q$ in a finite set $Q$. The probability that query $q$ is submitted at time $t$ is given by the probability distribution $F_Q \in \Delta(Q)$. The ad service then runs an auction to determine which ad to show in the given slot.

There are $N$ (potential) bidders, whom we also refer to as advertisers, indexed by $i = 1,\ldots,N$. Each bidder $i$ is characterized by a \emph{type} $\tau_i$ drawn at the beginning of each period from a finite set $\mathcal T_i$ according to a distribution $F_i \in \Delta(\mathcal T_i)$. We also fix functions $V_i : \mathcal T_i \times Q \to \mathbb R$ and $\mathit{CTR}_i : \mathcal T_i \times Q \to [0,1]$ such that, whenever bidder $i$'s type is $\tau_i \in \mathcal T_i$ and the shopper query is $q \in Q$, $i$'s \emph{value per click} is $V_i(\tau_i, q)$ and the ad \emph{click-through rate} is $\mathit{CTR}_i(\tau_i, q)$; these functions are in practice represented by matrices. 

In every period, each bidder $i$ simultaneously submits a \emph{bid} $b_i \in \mathbb R$ and a \emph{targeting clause} $c_i \subset Q$. The interpretation is that a targeting clause is the subset of possible shopper queries that bidder $i$ wants to bid on: the bidder participates in the auction only if the user's query $q$ is an element of $c_i$. The timing is as follows: first, bidder $i$ observes her type $\tau_i$; then, she places a bid $b_i$ and a targeting clause $c_i$; finally, the user query $q \in Q$ is realized. Thus, bidders \emph{cannot} tailor their bid to the specific user query in a given period. On the other hand, their bids can and will in practice depend upon their realized type. The winner of the auction is the bidder with the highest \emph{score}, i.e., the product of their bid and their click-through rate, provided they targeted the realized query. Formally, if the realized user query is $q$ and each bidder $i$ has type $\tau_i$, bids $b_i$, and targets $c_i$, then bidder $i$'s score is $s_i = b_i \cdot \mathit{CTR}_i(\tau_i, q)$ if $q \in c_i$ and $0$ otherwise; the winner is any $i \in \arg \max_{j \in N : q \in c_j} s_j$. Ties are broken randomly.

If the realized user query is $q$, bidder $i$ wins the auction, and the charged \emph{price per click} is $p$, then the average reward (payoff) to bidder $i$ with type $\tau_i$ is
$\mathit{CTR}_i(\tau_i, q) \cdot \left[ V_i(\tau_i, q) - p \right]$. Finally, if bidder $i$'s targeting clause $c_i$ does not include the realized $q$, or if it does but $b_i$ is not the winning bid, her payoff is 0. 

The price charged to the winner depends upon the auction rules; we consider different cases in \S \ref{sec:results}, so we defer the specifics until then. We also specify what bidders observe at the end of each period in \S \ref{sec:learning}, as that is a function of the learning algorithm under consideration. However, we maintain throughout that bidders \emph{only} observe their own reward, and not others' rewards or bids.

The concept of ``type'', as introduced by \citet{harsanyi1967games}, is key to model games with incomplete information: in every period, each bidder $i$ knows the distribution $F_j$ of her opponents $j$, but not their realized type. In one common interpretation, there is a population of potential bidders for every bidder role $i$, each with distinct value per click and click-through rate; in each period $t$, and for every bidder role $i$, a specific element of the corresponding population is drawn according to $F_i$. A symmetric environment is one where  the set of bidder types and their distribution are the same for every bidder; we use this in \S \ref{sec:results} for aggregate-level bidding behavior analysis. One alternative interpretation is that there is a single player (a firm) for every bidder role $i$; her type can represent the fact that the firm may be selling a range of goods, with per-period variations in values and click-through rates reflecting product-specific margins and conversion rates, perhaps due to cost variability, promotions, etc.\footnote{In either interpretation, formally, types are indices for the functions $V_i$ and $\mathit{CTR}_i$ with no intrinsic meaning.}

Types can also be used to model (indirectly) a \emph{random number of bidders}. Suppose that, for each bidder (or for a subset of bidders), we define a type $\tau_i^0$ who has zero value and click-through rate for every query. Then, in our learning algorithms, this bidder type will eventually stop bidding. Hence, out of $N$ potential bidders, only a subset may be active, the ones whose types are different from $\tau_i^0$. We assume type distributions are independent across bidders, although this could be relaxed as our learning algorithms do not rely on independence.

\emph{Standard auctions} are a special case of our model in which there is a single query: this can be captured by assuming that the set $Q$ of queries is a singleton, say $\{q^*\}$. Thus, a bidder's type $\tau_i$ fully characterizes her value for the object, and one may as well take $V_i$ to be the identity: $V_i(\tau_i, q^*) = \tau_i$. Traditional ``textbook'' auctions do not consider click-through rates; this can be captured by assuming that each function $\mathit{CTR}_i(\cdot)$ is identically equal to 1. However, pay-per-click auctions are common in advertising, so we consider them ``standard'' as well.

Targeting clauses provide flexibility in ad auctions. For instance, queries about USB chargers can vary in specificity and suggest differing shopper expertise. This results in varied values and click-through rates for advertisers, causing them to target different query types. For example, a wide-range product advertiser might target ``USB charger'' and ``USB type C charger'', while an Apple-specific advertiser might target ``iPhone charger''.

\raggedbottom
\subsection{Learning Algorithms}
\label{sec:learning}

The two main learning models we analyze are \texttt{Hedge} (also referred to as \texttt{Exponential Weights} or \texttt{Multiplicative Weights}) algorithm \citep{auer1995gambling,freund1997decision} and \texttt{EXP3-IX} \citep{kocak2014EXP3IX}. We fix a grid $B$ of possible bid values. To streamline notation, for each bidder $i$ and time period $t$, we let $a_{i,\tau_i,t}$ denote the pair $(b_i, c_i) \in B \times 2^Q \equiv A$ of bid and targeting clause chosen by bidder $i$ of type $\tau_i$ in period $t$. Relative to standard expositions of these algorithms \citep[e.g.]{lattimore2020bandit}, we add incomplete information and incorporate the draw of different queries in each iteration. 

Each bidder type learns from their own observations, not those of other types of the same bidder, to accommodate incomplete information. Learning is not conditioned on the unobserved realized query, reflecting the reality of online ad auctions.

\subsubsection{Model setup for auction environment}
We use the notation in Table \ref{tab:Notation-for-auction} to describe the auction environment and learning algorithms. Suppose each bidder $i$ bids $b_{i}$ on clause
$c_{i}$, and prices per query are $p=(p_{q})$. Denote by $c_{iq}$ the indicator function that equals 1 if and only if $q \in c_i$. Then bidder $i$'s
expected utility if she wins the auction for the (non-empty) set of queries $\mathcal Q \subseteq \{0,\ldots,Q-1\}$, given her value and CTR for each query $q \in \mathcal Q$ as $v_{iq}$ and $CTR_{i,q}$ respectively, is
\begin{equation}
\label{eq:utility}
EU_{i}(b_i,c_i,p)=\sum_{q \in \mathcal Q} F_Q(q) \cdot CTR_{iq}\cdot c_{iq}\cdot\left[v_{iq}-p_{q}\right],
\end{equation}
where the expectation is due to the random arrival of queries ($F_Q(q)$) and clicks ($CTR_{iq}$). In case of ties, for simplicity of implementation, we just divide the winner's surplus $v_{iq}-p_{q}$ by $N$, rather than (as we should)
the number of tying bidders. Losing bidders get 0 expected utility. Since the \texttt{EXP3-IX} algorithm requires rewards in $[0,1]$, we define
the \textbf{normalized reward} of bidder $i$ as
\begin{equation}
    \label{eq:reward-norm}
    r_{i}=\frac{EU_{i}(b_i,c_i,p)-(0-b_{\max})}{(v_{\max}-0)-(0-b_{\max})}=\frac{EU_{i}(b_i,c_i,p)+b_{\max}}{v_{\max}+b_{\max}}.
\end{equation}
\vskip -0.1in
\begin{table}[H]
\caption{\label{tab:Notation-for-auction}Notation for online ad auction}
\begin{centering}
\vskip 0.15in
\begin{tabular}{|c|c|}
\hline 
Object & Notation\tabularnewline
\hline 
\hline 
Bidders & $i=0,\ldots,N-1$\tabularnewline
\hline 
Time horizon & $T$\tabularnewline
\hline 
Bids & $b_{i}\in B \equiv \{0,...,b_{\max}\}$\tabularnewline
\hline 
Queries & $q\in\{0,...,Q-1\}$\tabularnewline
\hline 
Clause & $c_{i}=(c_{iq})_{q=0}^{Q-1}\in\{0,1\}^{Q}$\tabularnewline
\hline 
Value per bidder per query & $v_{iq}\in\{0,...,v_{\max}\}$\tabularnewline
\hline 
CTR per bidder per query & $CTR_{iq}\in[0,1]$\tabularnewline
\hline 
Normalized reward per bidder & $r_i\in[0,1]$\tabularnewline
\hline
\end{tabular}
\par\end{centering}
\end{table}

This is because the EU of a bidder cannot exceed the maximum value of the slot for any query if she were to get it for free, i.e., $v_{\max}-0$;
and it is always at least as large as getting a worthless slot and paying the maximum bid for it, i.e., $0-b_{\max}$. 

In our auction setting, we apply the \texttt{Hedge} and \texttt{EXP3-IX} algorithms as described below. Bidders independently use these to update their actions based on observed rewards. \texttt{Hedge} assumes bidders observe rewards for all possible actions, leading to faster learning but less realistic scenarios. In \texttt{Hedge}, $r_i$ refers to the time-$t$ normalized reward of bidder $i$'s type $\tau_i$ computed using Eq. \ref{eq:reward-norm}. 

\begin{algorithm}[H]
\begin{algorithmic}
\medskip
\REQUIRE parameters $\eta>0$, $N$, $T$
\REQUIRE action sets $A$ (the same for all bidders), query distribution $F_Q$
\REQUIRE type space $\mathcal T_i$ and distribution $F_i$, and functions $V_i, \mathit{CTR}_i$ for each $i = 1,\ldots,N$
\STATE for all $i$, $\tau_i \in \mathcal T_i$, and $a \in A$, set $w_{i,\tau_i,0}(a) = 0$
\FOR{$t = 1, \ldots, T$}
    \STATE draw $q \sim F_Q$
    \FOR[Actual play]{$i = 1,\ldots, N$}
        \STATE draw $\tau_i \sim F_i$
        \FOR[Softmax]{$b_i \in A$}
            \STATE set $p_{i,\tau_i,t-1}(b_i) = \frac{\exp(w_{i,\tau_i,t-1}(b_i)/\eta)}{\sum_{b_i' \in A}\exp(w_{i,\tau_i,t-1}(b_i')/\eta)}$
        \ENDFOR
        \STATE draw $a_{i,\tau_i,t} \sim p_{i,\tau_i,t-1}$
    \ENDFOR
    \FOR[Learning]{$i = 1,\ldots, N$}
        \FOR{$b_i \in A$}
            \STATE compute reward $r_i$ given $q$ and $\tau_i$, assuming $i$ bids $b_i$ and all $j \neq i$ bid $a_{j,t}$
            \STATE set $w_{i,\tau_i,t}(b_i) = w_{i,\tau_i,t-1}(b_i) + r_i$
            \FOR{$\tau_i' \in \mathcal T_i \setminus \{\tau_i\}$}
                \STATE set $w_{i,\tau_i',t}(b_i) = w_{i,\tau_i',t-1}(b_i)$
            \ENDFOR
        \ENDFOR
    \ENDFOR
\ENDFOR
\end{algorithmic}
\caption{Hedge for Auctions}
\label{alg:hedge}
\end{algorithm}

\texttt{EXP3-IX} assumes observation of the chosen action's reward, providing a slower but more realistic learning process. In \texttt{EXP3-IX}, the quantities $l_{i,\tau_i,t}$ are ``losses'' incurred at time $t$ by bidder $i$ of type $\tau_i$; that is, $\ell_{i, \tau_i, t} = 1 - r_{i}$. We chose \texttt{EXP3-IX} over \texttt{EXP3} due to its tighter bounds on realized regret, despite \texttt{EXP3}'s good expected regret guarantees ( \citep[Chap.12]{lattimore2020bandit}).
Although \texttt{Hedge} is more computationally efficient and less dispersed, a deeper investigation of these algorithms' distributional characteristics is suggested. 

\begin{algorithm}[H]
\begin{algorithmic}
\medskip
\REQUIRE parameters $\eta>0$, $\gamma>0$, $N$, $T$
\REQUIRE action sets $A$ (the same for all bidders), query distribution $F_Q$
\REQUIRE type space $\mathcal T_i$ and distribution $F_i$, and functions $V_i, \mathit{CTR}_i$ for each $i = 1,\ldots,N$ \\
for all $i$ and $a \in A$, set $l_{i,\tau_i,0}(a) = 0$
\FOR{$t = 1, \ldots, T$}
    \STATE draw $q \sim F_Q$
    \FOR[Play]{$i = 1,\ldots, N$}
        \STATE draw $\tau_i \sim F_i$
        \FOR[Softmax]{$a_i \in A$}
            \STATE set $p_{i,\tau_i,t}(a_i) = \frac{\exp(-\eta \ l_{i,\tau_i,t-1}(a_i))}{\sum_{a_i' \in A}\exp(-\eta \  l_{i,\tau_i,t-1}(a_i'))}$
        \ENDFOR
        \STATE draw $a_{i,\tau_i,t} \sim p_{i,\tau_i,t}$
    \ENDFOR
    \FOR[Learning]{$i = 1,\ldots, N$}
        \STATE compute \emph{normalized} reward $r_{i,t} \in [0,1]$ given $q$ and $\tau_i$ under the actual bids $a_{i,\tau_i,t}$ and set 
        \STATE {\small $l_{i,\tau_i,t}(a_{i,\tau_i,t}) = l_{i,\tau_i,t-1}(a_{i,\tau_i,t}) + \frac{1-r_{i,t}}{p_{i,\tau_i,t}(a_{i,\tau_i,t})+\gamma}$}
        \FOR{$a_i' \in A \setminus \{a_{i,\tau_i,t}\}$ and $\tau_i' \in \mathcal T_i \setminus \{\tau_i\}$}
            \STATE set $l_{i,\tau_i',t}(a_i') = l_{i,\tau_i',t-1}(a_i')$
        \ENDFOR
    \ENDFOR
\ENDFOR
\end{algorithmic}
\caption{\texttt{EXP3-IX} for Auctions}
\label{alg:EXP3-IX}
\end{algorithm}

\subsection{Convergence to equilibrium}
\label{sec:coarseBCE}

Since both \texttt{Hedge} and \texttt{EXP3-IX} ensure that each type will have vanishing regret in the limit as $T \to \infty$, Lemma 10 in \citet{hartline2015coarseBCE} implies that the resulting dynamics will converge to a version of correlated equilibrium for games with incomplete information. We now formally define this equilibrium notion for the class of games we consider. 

For every bidder $i$, let $\mathbf s_i: \mathcal T_i \to B \times \{0,1\}^Q$ be a \emph{strategy} for bidder $i$: for each type $\tau_i \in \mathcal T_i$, a strategy specifies a bid and a targeting clause: $\mathbf s_i(\tau_i) = (b_i,c_i)$. Let $\mathbf S_i$ be the set of all  such strategies for $i$, and define the Cartesian product sets $\mathbf S = \prod_i \mathbf S_i$ and $\mathbf S_{-i} = \prod_{j \neq i} \mathbf S_j$ as usual. For $\mathbf s \in \mathbf S$ and $\tau \in \prod_i \mathcal T_i$, by convention $\mathbf s(\tau) = \left(\mathbf s_i(\tau_i) \right)_{i \in N}$; similarly,  $\mathbf s_{-i}(\tau_{-i}) = \left(\mathbf s_j(\tau_j) \right)_{j \neq i}$.

To describe the outcome of the auction, for every query $q$, let $\mathbf p_q : (B \times \{0,1\}^Q)^N \to \mathbb R$ denote the pricing rule of the auction, which associates the price charged to the winner with any vector of bids and targeting clauses. Furthermore, for each $i = 0,\ldots,N-1$ and $q \in Q$, let $\mathbf w_{iq} : (B \times \{0,1\}^Q)^N \to [0,1]$ denote the probability that bidder $i$ wins the auction for query $q$ given the bids and clauses submitted by all players. We assume that, for every query $q$ and tuple $(b,c) \in (B \times \{0,1\}^Q)^N$, $\sum_i \mathbf w_{iq}(b,c) = 1$  and $\mathbf w_{iq}(b,c)>0$ only if $c_{iq} = 1$; that is, in order to win the auction for a query, the bidder must have included it in her targeting clause.

We can then define the \emph{payoff} of bidder $i$'s type $\tau_i$, given a tuple $(b,c) \in (B \times \{0,1\}^Q)^N$ of bids and targeting clauses for every player, as
\[
U_i(b,c; \tau_i) = \sum_{q=0}^{Q-1} F_Q(q) \cdot CTR_{iq}(\tau_i)\cdot \mathbf w_{iq}(b,c) \cdot\left[V_{iq}(\tau_i)-\mathbf p_{q}(b,c)\right].
\]
With these definitions, a probability distribution $\sigma \in \Delta(\mathbf S)$ is a \textbf{coarse Bayes Correlated Equilibrium} (coarse BCE) if, for every $i = 0,\ldots,N-1$, $(b_i,c_i) \in B \times \{0,1\}^Q$, and $\tau_i \in \mathcal T_i$,
\[
\mathrm E_{\mathbf s \sim \sigma, \tau_j \sim F_j: j \neq i} U_i(\mathbf s(\tau); \tau_i) \geq \mathrm E_{\mathbf s \sim \sigma, \tau_j \sim F_j: j \neq i} U_i\left( (b_i, c_i), \mathbf s_{-i}(\tau_{-i}); \tau_i\right).
\]
That is, for each type $\tau_i$, there is no single action $(b_i,c_i)$ that improves upon her payoff if she conforms to the profile $\sigma$. This differs from the notion of a (non-coarse) Bayes Correlated Equilibrium \citep{bergemann2016bayes} in that it does not rule out more sophisticated deviations, in which type $\tau_i$ chooses different actions $b_i,c_i$ for each recommended action $\mathbf s_i(\tau_i)$ in the support of the marginal of the equilibrium distribution $\sigma$ on $i$'s strategy space $\mathbf S_i$.

\raggedbottom
\section{Results}
\label{sec:results}
In this section, we provide our results. T he code implementing the learning algorithms described in \S\ref{sec:learning} is shared in our submission and will be available on GitHub. 

\subsection{Standard auctions}
\label{sec:results:textbook}

We first consider a standard, symmetric auction environment with a single query, so $Q = \{q^*\}$, and pay-per-impression pricing, so all CTRs are equal to 1. We consider both a small auction with $N = 2$ bidders, and a more realistic one with $N=10$ bidders. Values are uniformly distributed on a grid $B$ in the interval $[0,1]$. Formally, given the grid $B$, for all $i$, we let $\mathcal T_i = B$, $V_i(b, q^*) = b$, $\mathit{CTR}_i(b, q^*) = 1$, and $F_i(b) = \frac{1}{|B|}$ for all $b \in B$. Furthermore, we allow all bids $b_i$ on the grid $B$.

In a second-price auction, it is a dominant strategy for every bidder to bid their value. In a first-price auction, with $N$ bidders, a continuum of values (uniformly distributed on $[0,1]$) and bids, the unique symmetric Bayesian Nash equilibrium is for each bidder to bid a fraction $\frac{N-1}{N}$ of their value. Furthermore, Myerson's revenue equivalence applies, so the expected revenue for the advertising service is the same under both formats, namely $\frac{N-1}{N+1}$. We want to compare these theoretical predictions with the output of our learning algorithms with a fine enough grid $B$, and a long enough horizon $T$.

We choose $B = \{\frac{i}{20} : i = 0,\ldots, 20\}$ and $T = 1,000,000$ for \verb|EXP3-IX|. We also choose the tuning parameters optimally, as described in \citet{lattimore2020bandit} (Theorem 12.1, Chapter 12).  Table \ref{tab:standard-EXP3IX} displays revenues for the advertising service in first- and second-price auctions under \verb|EXP3-IX|, averaged over the last $10\%$ of the learning period, i.e., $100,000$ iterations, and 5 different runs of the algorithm, as well as the standard deviation of revenues across different runs\footnote{Throughout the paper, we draw a fixed sequence of type realizations for all bidders. This ensures that the only randomness is from the algorithms' choice of action.}.
%
%
\begin{table}[ht]
    \caption{Standard single-item auction, \texttt{EXP3-IX}}
    \label{tab:standard-EXP3IX}
    \vskip 0.1in
    \centering
    \begin{tabular}{|l|c|c|}
    \hline
    Auction format & Mean Revenue & Std. Dev.  \\ \hline
    First price, $N=2$ & $0.3474$ & $0.0205$ \\
    Second price, $N=2$ & $0.3346$ & $0.0016$ \\ \hline
    First price, $N=10$ & $0.780$ & $0.0095$ \\
    Second price, $N=10$ & $0.7723$ & $0.0171$ \\ \hline
    \end{tabular}
\end{table}

Table \ref{tab:standard-Hedge} reports the results for \verb|Hedge|. We choose $T = 400,000$ and $\eta = 0.02$. We averaged over 5 different runs of the algorithm and over the last $40,000$ time periods.
\begin{table}[H]
    \caption{Standard single-item auction, \texttt{Hedge}.}
    \label{tab:standard-Hedge}
    \vskip 0.1in
    \centering
    \begin{tabular}{|l|c|c|}
    \hline
    Auction format & Mean Revenue & Std. Dev.  \\ \hline
    First price, $N=2$ & $0.3201$ & $0.0018$ \\
    Second price, $N=2$ & $0.3256$ & $0.0018$ \\ \hline
    First price, $N=10$ & $0.8305$ & $0.0011$ \\
    Second price, $N=10$ & $0.8334$ & $0.0008$ \\ \hline
    \end{tabular}
\end{table}

The key take-away is that expected revenues are close to the theoretical value $\frac13$ for $N=2$ under both $\texttt{Hedge}$ and $\texttt{Exp3-IX}$. With $N=10$, $\texttt{Hedge}$ again approximates the theoretical value $0.\overline{81}$, while $\texttt{Exp3-IX}$ is not as close. Revenues do not vary much across different runs. Even when $N=2$, \verb|EXP3-IX| required a considerably larger number of periods to achieve similar results
as \verb|Hedge|. 

\subsection{Soft floors}
\label{sec:results:SFRP}

Soft floors, also known as soft-floor reserve prices, are commonly used in online advertising; to the best of our knowledge, \citet{zeithammer2019soft} provides the most exhaustive formal analysis to date. 

This pricing mechanism works as follows; we consider the case of equal click-through rates for simplicity. Let $s > 0$ denote the soft-floor, and $b_{(1)}$ and $b_{(2)}$ be the first- and, respectively, second-highest bids. If $b_{(2)} \geq s$, then the price charged is $p = b_{(2)}$, as in a standard second-price auction. If $b_{(1)} \geq s > b_{(2)}$, then $p = s$, as if $s$ was a standard reserve price (``hard floor''). Crucially, if $s > b_{(1)}$, then the high bidder still wins the auction (on the contrary, with a standard reserve price, the seller would keep the object) and $p = b_{(1)}$. 

\citet{zeithammer2019soft} shows that, in a symmetric auction for a single object, bid functions are monotonic. As a consequence, the revenue equivalence theorem \citep{myerson1981optimal} applies\footnote{Intuitively, under different pricing rules, bidding behavior is also different, in a way that exactly offsets differences in the way prices are computed.}, and introducing soft-floors in second-price auctions do not affect either the final allocation or the advertising service's revenues. He then demonstrates by way of examples that, with asymmetric bid distributions, revenues in a second-price auction with a soft-floor can be either higher or lower than in a standard second-price auction.

As noted in \S\ref{sec:intro}, our contribution here is twofold. First, we demonstrate through simulations that, with multi-query targeting, different auction formats---and in particular auctions with a soft floor---\emph{can} yield different revenues, even if bidder types are drawn from the same distribution. Second, restricting attention to single-query auctions, we consider a collection of asymmetric bid distributions that complement those analyzed by \citet{zeithammer2019soft}, and do not allow for analytical equilibrium solutions. For such environments, we demonstrate how a second-price auction with a suitably chosen reserve price yields higher revenue than a soft-floor auction. 

For completeness, we first consider the symmetric, single-query environment of \S\ref{sec:results:textbook} and simulate an auction with a soft-floor equal to $s = 0.5$. Consistently with \citet{zeithammer2019soft}, we find that soft-floors have virtually no impact on revenues.\footnote{Results are as follows (compare with Tables \ref{tab:standard-EXP3IX} and \ref{tab:standard-Hedge}): \texttt{Hedge} with $N=2$ bidders yields average revenues equal to $0.324$ (standard deviation $0.0024$); \texttt{Hedge} with $N=10$ yields $0.834$ ($7e-6$); \texttt{EXP3-IX} with $N=2$ yields $0.377$ ($0.032$); \texttt{EXP3-IX} with $N=10$ yields $0.774$ ($0.0176$).}

Now turn to a multi-query environment, which, as noted, is beyond the scope of \citet{zeithammer2019soft}. Our objective is to demonstrate the \emph{possibility} of raising revenues in symmetric environments through soft floor; to do so, we analyze the following parameterized example.

We assume $N = 3$ bidders, all with the same set of possible types $\mathcal T_1 = \mathcal T_2 = \mathcal T_3 =  \{1,2,3\}$. There are two queries, so $Q = \{1, 2\}$. Values and click-through rates for all bidders are as described in Table \ref{tab:SFRP}. Thus, for example, $V_i(2,1) = 0.25$ and $\mathit{CTR}_i(1, 2) = 0.1$. Both queries are equally likely, and all types are also equally likely. We let $B = \{\frac{i}{20} \: : \: i = 0,\ldots,20\}$ and $T = 1,000,000$.\footnote{We also ran these simulations with higher values of $T$, and similar patterns emerge.}
\begin{table}[ht]
\caption{Values and Click-Through Rates for all bidders.}
\vskip 0.1in
\label{tab:SFRP}
\centering
\begin{tabular}{|c|c|c|c|c|}
\hline
$\tau_i$ & $V_i(\tau_i, 1)$ & $\mathit{CTR}_i(\tau_i, 1)$ &  $V_i(\tau_i, 2)$ & $\mathit{CTR}_i(\tau_i, 2)$ \\ \hline
1 & 0.5 & 0.3 & 0.25 & 0.1 \\
2 & 0.25 & 0.1 & 1 & 0.1 \\
3 & 0.25 & 0.1 & 1 & 0.2 \\ \hline
\end{tabular}
\end{table}

Table \ref{tab:SFRP-results-EXP3IX-ming} reports expected revenues per impression (where expectations are taken over search queries, bidder types, and click-through rates) for \verb|EXP3-IX|, averaged over 5 runs. Table \ref{tab:SFRP-results-Hedge-ming} reproduces the results for \verb|Hedge|, with $T = 400,000$.

\begin{table}[ht]
    \caption{Multi-query auctions and soft-floors, \texttt{EXP3-IX} output for $T = 1\mathrm M$, averaged over 5 runs.}
    \vskip 0.1in
    \label{tab:SFRP-results-EXP3IX-ming}
    \centering
    \begin{tabular}{|l|c|c|}
    \hline
    Auction format & Mean Revenue & Std. Dev.  \\ \hline
    First price &  $0.0830$ & $0.0007$ \\
    Second price & $0.0509$ & $0.0008$ \\ 
    soft-floor $s=0.65$ & $0.0813$ & $0.0007$ \\ \hline
    \end{tabular}
\end{table}

\begin{table}[ht]
    \caption{Multi-query auctions and soft-floors, \texttt{Hedge} output for $T = 400 \mathrm K$, averaged over 5 runs.}
    \vskip 0.1in
    \label{tab:SFRP-results-Hedge-ming}
    \centering
    \begin{tabular}{|l|c|c|}
    \hline
    Auction format & Mean Revenue & Std. Dev.  \\ \hline
    First price &  $0.0691$ & $0.0016$ \\
    Second price & $0.0857$ & $0.0001$ \\ 
    soft-floor $s=0.65$ & $0.0741$ & $0.0061$ \\ \hline
    \end{tabular}
\end{table}

Our first key finding is that, as anticipated, the three auction formats yield different expected revenues.  Soft-floor reserve prices can impact revenues; thus, our simulations provide some support to this common industry practice. Under \texttt{Hedge}, revenues are higher in soft-floor reserve price auctions than in first-price auctions, but second-price auctions perform best. With \texttt{EXP3-IX}, second-price auctions do not fare as well; the highest revenues come from first-price auctions, with soft-floor pricing behind. 

\begin{figure}[ht]
    \centering
    \includegraphics[scale=0.37]{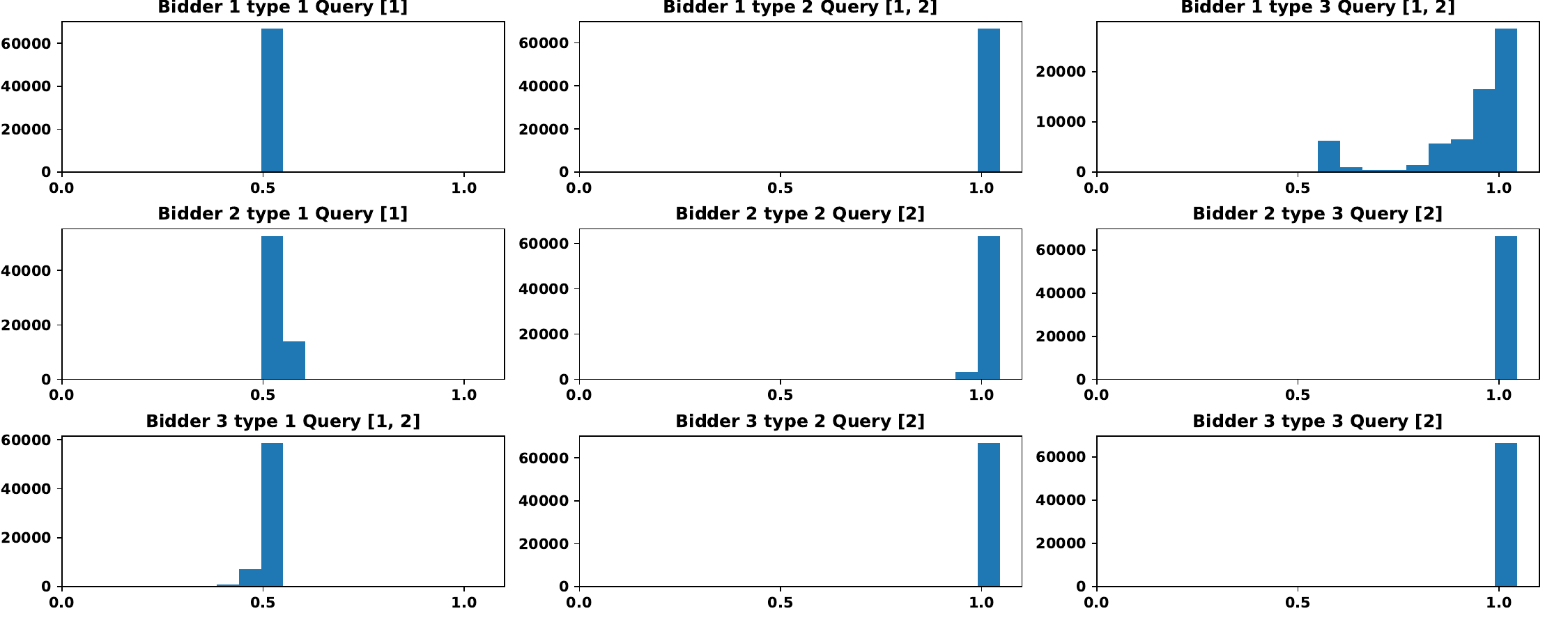}
    \caption{Bids in second-price auction under \texttt{Hedge}}
    \label{fig:hedge-SP}
\end{figure}

A second key finding is that \texttt{EXP3-IX} results do not align with \texttt{Hedge} even when run for longer periods. Each panel in Figure \ref{fig:hedge-SP} shows the frequency of bids in the final $40,000$ periods of the learning algorithm, summed over 5 runs, divided by type and targeting clause (i.e., queries actually targeted). The main take-away point is that every type eventually learns to choose a specific targeting clause \emph{and} for the most part also places fairly concentrated bids. 

\begin{figure}[ht]
    \centering
    \includegraphics[scale=0.37]{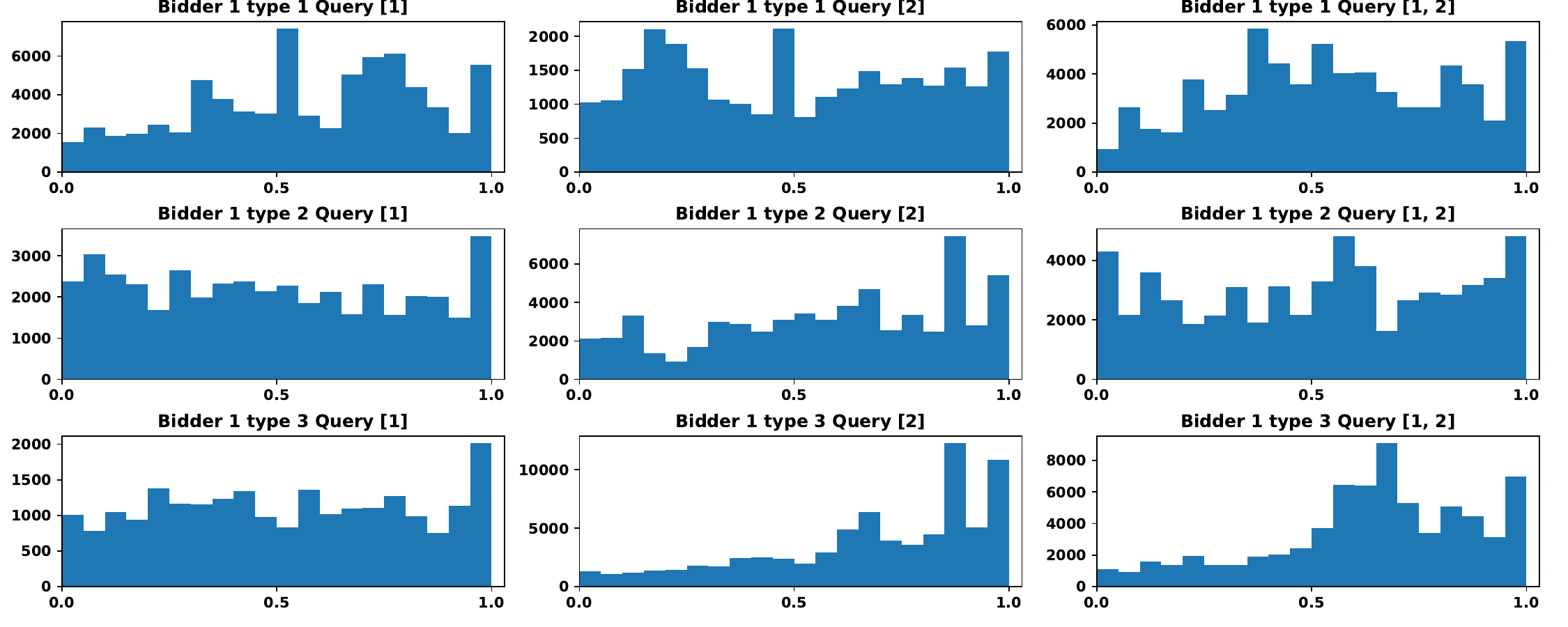}
    \caption{Bidder 1's bids in second-price auction, \texttt{EXP3-IX}}
    \label{fig:EXP3IX-SP}
\end{figure}

Now compare with Figure \ref{fig:EXP3IX-SP}, which summarizes predicted bids for bidder 1 under \texttt{EXP3-IX}, with Figure \ref{fig:hedge-SP} above (again we sum over 5 runs and only look at the last $100,000$ periods).

Type 1 occasionally bids on one or both queries, showing little convergence even after one million runs. \texttt{EXP3-IX} requires more experimentation, as it only learns from played actions, while \texttt{Hedge} converges faster but assumes learning about unplayed actions. This supports the idea that leveraging knowledge of the pricing mechanism enhances performance. \texttt{Hedge} outperforms due to its reliance on full reward information. We suggest that incorporating heuristics into \texttt{EXP3-IX} may partially address this imbalance; we leave this to future work.

\subsection{Soft-floor vs standard reserve price with a single query}
\label{sfrp_vs_rp}

We now restrict attention to single-query auctions, and explore the implications of bidder asymmetry. Recall that standard reserve prices in second-price auctions work as follows. Again, let $b_{(1)}$ and $b_{(2)}$ denote the first- and second-highest bids, and assume that click-through rates are the same for all bidders for simplicity. Given a reserve price $r > 0$, if $b_{(2)} \geq r$ then the high bidder receives the slot and pays $b_{(2)}$; if $b_{(1)} \geq r > b_{(2)}$, the high bidder receives the slot but pays $r$ (the reserve provides ``price support''). Finally, if $r > b_{(1)}$, the slot is not sold. As noted above, this is the key difference between standard and soft floors.

A common concern in ad auctions, which motivated the introduction of soft floors, is the possible presence of dominant bidders in a market---bidders whose valuation is substantially higher than that of other bidders. The key question is what is meant by ``substantially higher.'' 

\citet{zeithammer2019soft} considers two qualitatively different parametric illustrations. In the first, one bidder has values drawn from the $[0,1]$, and the other from $[0,M]$, with $M \geq 0$. The latter bidder is the ``stronger'' participant. Crucially, this stronger bidder is only present in the auction with probability $\alpha$. In this case (see Fig. 5 in \citealp{zeithammer2019soft}), a soft floor \emph{may} yield higher revenues than a standard reserve price. The second case is that of a randomly appearing strong bidder, with values drawn from an interval $[L,M]$ with $L \geq 1$, facing one or more ``regular'' bidders with values in $[0,1]$. Here, a closed-form solution is not possible in all cases. Proposition 4 shows that, if the soft floor $s$ is \emph{lower} than the highest bid of the regular bidders in a first-price auction, the auctioneer would obtain higher revenues by using the vale $s$ as standard reserve price. However, the model is analytically intractable if $s$ is above the highest bid of regular bidders in a first-price auction. 

We thus perform a variety of simulations that include such intractable scenarios. In our first experiment, we consider two regular bidders with i.i.d., equally likely values of $0.2, 0.4, 0.6, 0.8$, and $1.0$.

Also, two strong bidders with value 2 participate with i.i.d. probability of 0.5. Consider soft-floors from 0 (which corresponds to a simple second-price auction) to 2 (effectively, a first-price auction with a reserve of 2). For each parameter value, we run 500,000 simulation periods and compute average revenues in the final 50,000 periods. We report the average of the results over 5 repetitions of the simulation. 

We then repeat the entire set of simulations, this time using a standard reserve price rather than a soft-floor reserve price. We consider only the values in the support of the regular bidders’ value distribution, plus $1.8$ and $2.0$: this is because a reserve price of $0.0$ is the same as a soft-floor of $0.0$, and reserve price between $1.2$ and $1.6$ exclude the low bidders, just like a reserve price equal to $1.8$, but provides a lower price support than $1.8$. Therefore, reserve prices strictly between $1.0$ and $1.8$ always yield strictly lower revenue than either $1.8$ or some reserve price between $0$ and $1$; for this reason, we do not include them in our graphs and tables.

The results in Table \ref{tab:sfrp-vs-rp-first} show that, even without optimizing hard floors, there is no benefit to soft-floors, whether high or low; this is true, even considering soft floors that are not covered by the analysis in \citet{zeithammer2019soft}.\footnote{In a first-price auction without a reserve price, no regular bidder would bid above $0.5$. Even with a higher reserve price, no regular bidder would bid above $1.0$. Thus, soft floors above $1.0$ are not covered by the results in \citet{zeithammer2019soft}.} In fact, soft-floors lead to lower revenues than standard reserve prices. One feature of the above parameterization is that, since the stronger bidders have a high valuation and at least one of them appears with high probability $(0.75)$, it is optimal for the seller to  target that bidder only, by fixing a high reserve price of $1.8$. This means that the regular bidders are excluded from participation. This may be undesirable even if it does maximize ad revenues. Yet even without completely shutting the low-valuation bidders out of the market, a moderate standard reserve price ($0.6$ in this case) yields higher revenues than any soft-floor, or no floor.

\begin{table}[h]
\centering
\caption{Ad revenue: soft-floor reserve price (SFRP) vs standard reserve price (RP). We exclude reserve prices between $1.0$ and $1.8$, as they yield lower revenue than reserve prices at $1.8$ or below $1.0$. RP=$0$ is the same as SFRP=$0$.}
\vskip 0.1in
\begin{tabular}{|c|c|c|c|c|}
\hline
Floor & Revenue (SFRP) & Stdev & Revenue (RP) & Stdev \\\hline
0.  & 0.965 & 0.0001  &         &         \\
0.2 & 0.985 & 0.0001  & 0.975   & 0.0002  \\
0.4 & 0.96  & 0.0001  & 0.9894  & 0.0003  \\
0.6 & 0.893 & 0.00006 & 0.9979  & 0.0001  \\
0.8 & 0.84  & 0.162   & 0.991   & 0.0002  \\
1   & 0.766 & 0.001   & 1.0053  & 0.0002  \\
1.2 & 0.836 & 0.056   &         &         \\
1.4 & 0.839 & 0.139   &         &         \\
1.6 & 0.861 & 0.064   &         &         \\
1.8 & 0.813 & 0.045   & 1.3656  & 0.0005  \\
2   & 0.836 & 0.113   & 0.0464  & 0.002   \\
\hline
\end{tabular}
\label{tab:sfrp-vs-rp-first}
\end{table}

\paragraph{Exploring different value distributions}
The caveat is that the preceding results are based on a specific distribution of values for the regular bidders---a uniform distribution. To explore the robustness of the conclusions, we re-run all simulations for a sample of discretized Beta distributions with different parameter values. These are chosen to represent right-skewed, left-skewed, and centered but differently concentrated value distributions. We also re-ran all simulations with standard reserve prices instead of soft-floors, as as in the preceding section.

\begin{figure}
    \begin{subfigure}{\textwidth}
        \centering
        \includegraphics[scale=0.5]{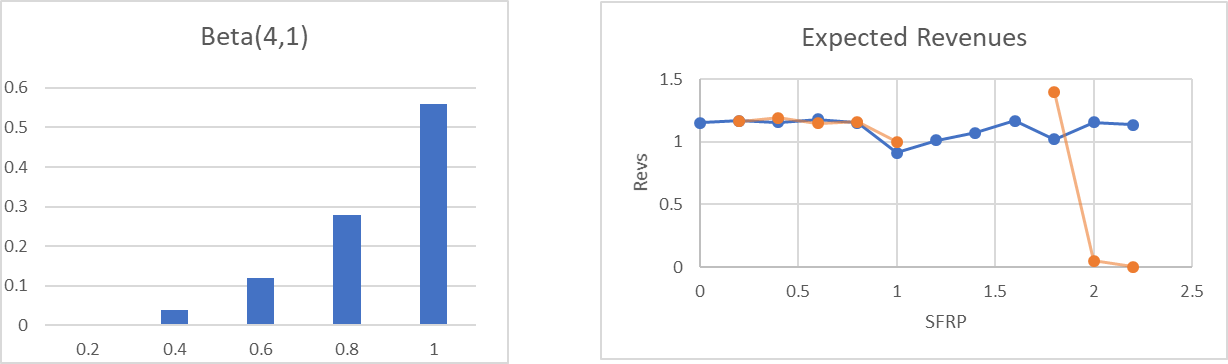}
    \end{subfigure}
    \vspace{0.06in} 
    \begin{subfigure}{\textwidth}
        \centering
        \includegraphics[scale=0.5]{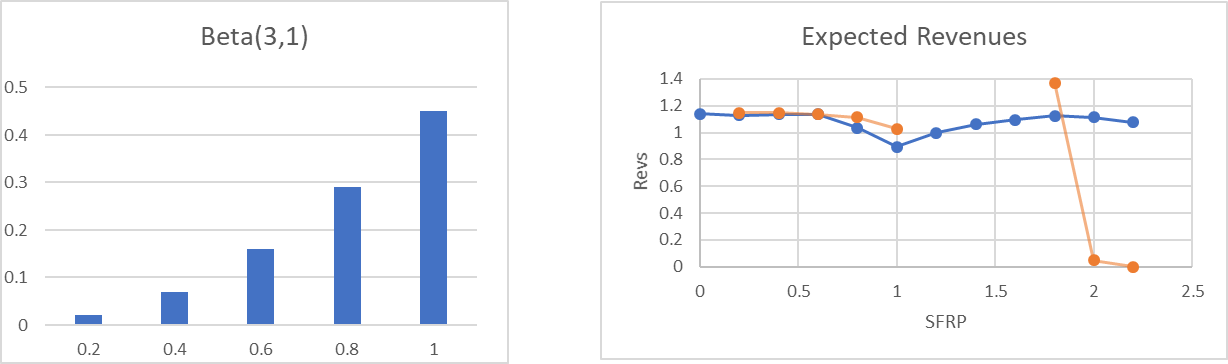}
    \end{subfigure}
    \vspace{0.06in} 
    \begin{subfigure}{\textwidth}
    \centering
        \includegraphics[scale=0.5]{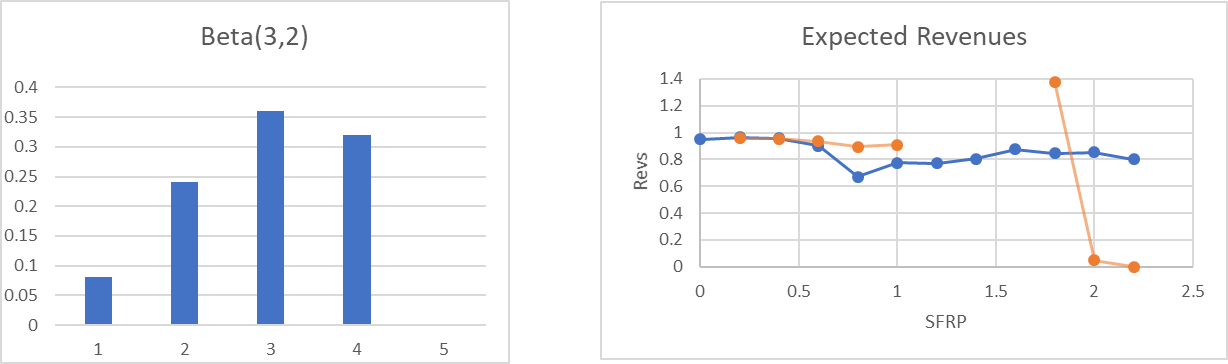}
    \end{subfigure}
    \caption{Expected revenue for soft-floor reserve price (SFRP, blue line) and standard reserve price (RP, orange line) under right-skewed value distributions. We exclude reserve prices between $1.0$ and $1.8$, as they yield lower revenue than reserve prices at $1.8$ or below $1.0$. RP=$0$ is the same as SFRP=$0$.}
    \label{right_skewed}
\end{figure}

\begin{figure}
    \begin{subfigure}{\textwidth}
        \centering
        \includegraphics[scale=0.5]{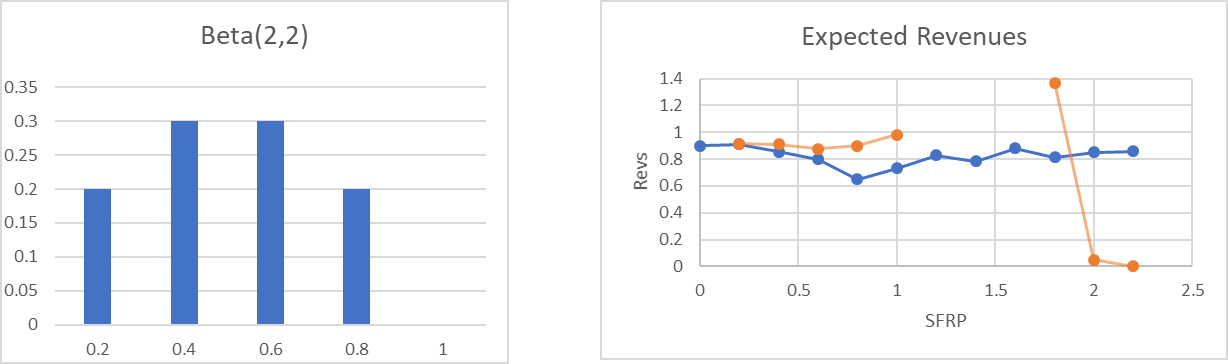}
    \end{subfigure}
    \vspace{0.06in} 
    \begin{subfigure}{\textwidth}
        \centering
        \includegraphics[scale=0.5]{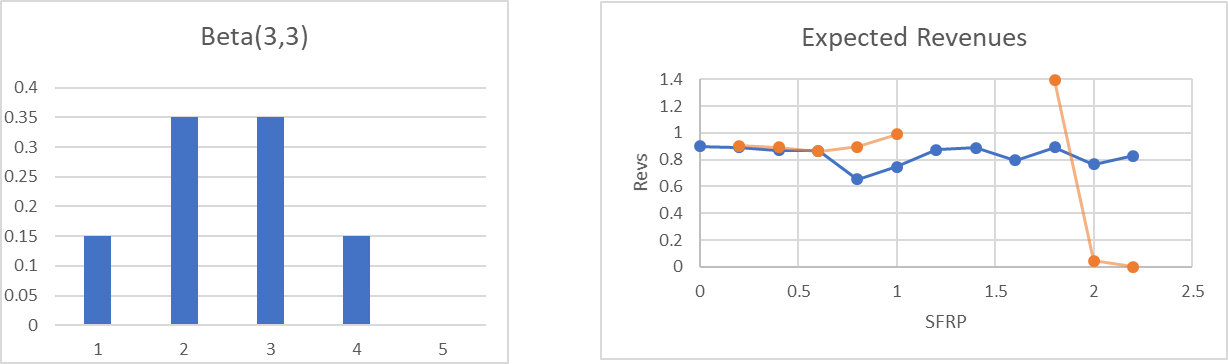}
    \end{subfigure}
    \vspace{0.06in} 
    \begin{subfigure}{\textwidth}
        \centering
        \includegraphics[scale=0.5]{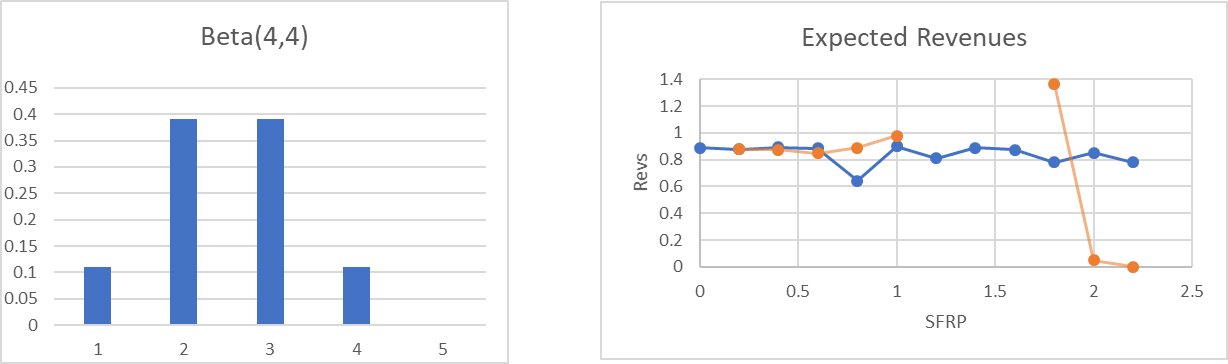}
    \end{subfigure}
    \caption{Expected revenue for soft-floor reserve price (SFRP, blue line) and standard reserve price (RP, orange line) under symmetric value distributions.}
    \label{symmetric}
\end{figure}

\begin{figure}
    \begin{subfigure}{\textwidth}
    \centering
        \includegraphics[scale=0.5]{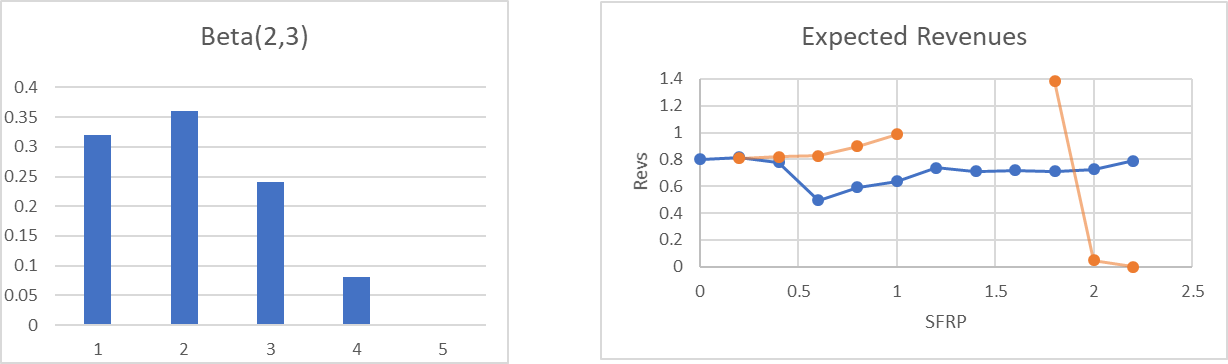}
    \end{subfigure}
    \vspace{0.06in} 
    \begin{subfigure}{\textwidth}
    \centering
        \includegraphics[scale=0.5]{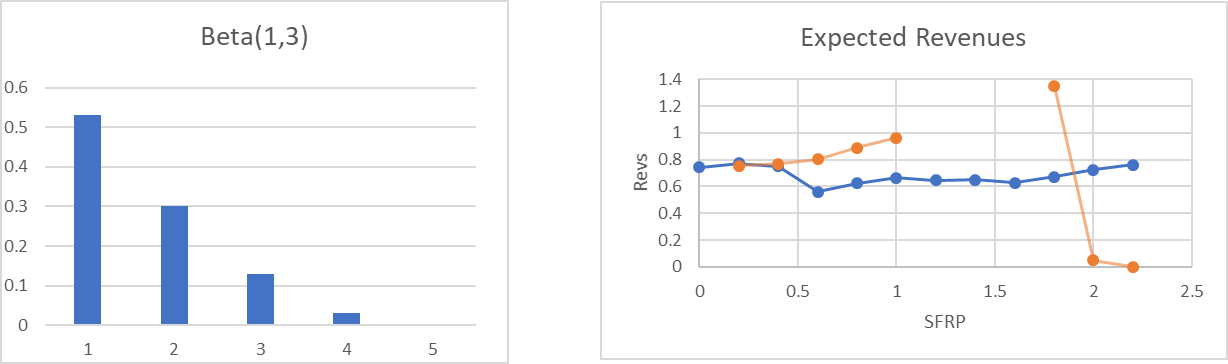}
    \end{subfigure}
    \vspace{0.06in} 
    \begin{subfigure}{\textwidth}
    \centering
        \includegraphics[scale=0.5]{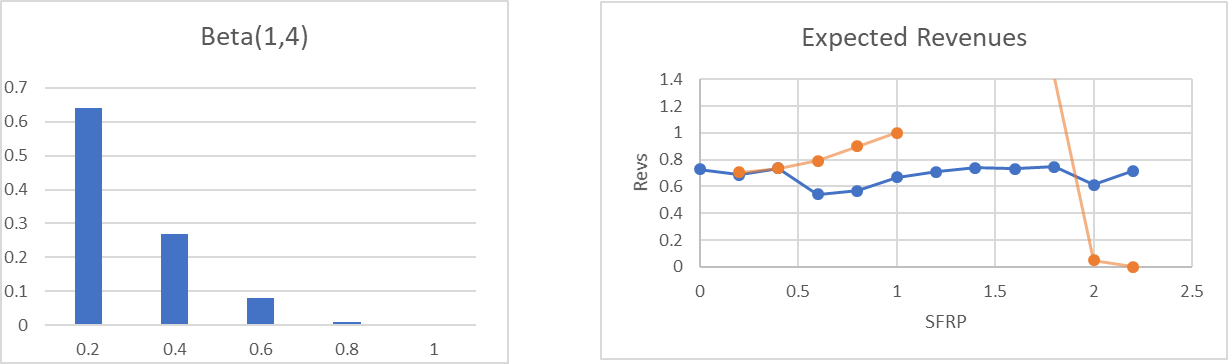}
    \end{subfigure}
    \caption{Expected revenue for soft-floor reserve price (SFRP, blue line) and standard reserve price (RP, orange line) under left-skewed value distributions.}
    \label{left_skewed}
\end{figure}

For each parameterization, the graphs below (Figures \ref{right_skewed}--\ref{left_skewed}) depict the value distribution (left panel) and the corresponding expected revenue for soft-floors between $0$  and $2.2$ (in blue), and standard reserve prices from $0$ to $1.0$, plus $1.8$ and $2.0$ (orange).
The results are qualitatively similar: soft-floors, whether low (in the range of values of the regular bidders) or high (above the highest value of regular bidders) do not lift revenues beyond those in a second-price auction. Furthermore, an optimally chosen reserve price does better than any soft-floor. Finally, even restricting attention to standard reserve prices between $0$ and $1$, one can still extract higher revenues than with a soft-floor. One additional finding is that noise is roughly increasing in the magnitude of the soft-floor, reflecting a more complex auction environment for bidders to learn.

\subsection{Inferring values and bid shading}
\label{sec:results:bid-shading-prod}

Next, we demonstrate how our approach can be used as a step in an inference procedure aimed at determining the distribution of bidders' values, which is not directly observable, from the observed distribution of bids. In \S\ref{evals}, to validate our approach, we start from randomly generated values, then simulate bids, and use them in an iterative procedure to retrieve values from bid. Then, in  \S\ref{prod}, we leverage our simulator in a production environment. This application involves inferring the bidder's value distributions for two scenarios: low and high traffic shopper queries. We utilize \texttt{Hedge} for the inference procedure, due to the bid dispersion observed with \texttt{EXP3-IX} (\S\ref{sec:results:SFRP}).

\subsubsection{Validation with randomly generated values}
\label{evals}

For expository purposes, to reduce the number of cases to consider, we consider auctions with a single query ($Q = 1$) and set all bidders' click-through rate to 1, though our approach is by no means limited to this case. This implies that types coincide with values for the query. We also assume bidders are symmetric, i.e. their types are drawn from the same distribution.

At a broad level, our analysis proceeds as follows. First, we choose different type (i.e. value) distributions and pricing rules, described below. Second, for each such type distribution and pricing rule, we run our auction simulation to generate bids, and compute the distribution of bids in the last 10\% of the learning periods. Henceforth, we call this the \emph{observed} distribution of bids, because it is a stand-in for bid data one might in principle have access to (whereas values are not directly observable).
Finally, we take as input the observed distribution of bids thus obtained, and infer the value distribution through an iterative procedure. 
We initialize this procedure by setting the \emph{inferred} value distribution equal to the \emph{observed} \underline{bid} distribution. We then run the learning algorithm and derive a \emph{predicted} bid distribution. In computing rewards, we need to select a pricing rule. 

The next step involves adjusting inferred values for each percentile. To do so, we use the following heuristic. Suppose that, for a given percentile, the currently inferred value is $v$, the observed bid is $b^o$, and the predicted bid is $b^p$. This means that the predicted extent of bid shading (reducing one's bid below one's value) is $\sigma = \frac{b^p}{v}$. We then update $v$ according to
\begin{equation}
\label{eq:inference}
v \leftarrow v + \alpha \left(  \frac{b^o}{\sigma} - v \right).
\end{equation}
Intuitively, a bidder with value $\frac{b^o}{\sigma}$ who bids by applying a shading factor of $\sigma$ will bid exactly $b^o$, the actually observed bid for this quantile. We then adjust the inferred $v$ in the direction of $\frac{b^o}{\sigma}$, but apply a learning rate adjustment $\alpha$ to reduce overshooting. We then use the inferred values thus obtained as input for the next iteration of the inference procedure. To compensate for randomness in the learning algorithm, we average predicted bids over multiple runs.

For the present paper, we generated three possible ``true'' value distributions: uniform on $[0,3]$, ``right-skewed'', and ``left-skewed.'' The latter two are, respectively, a discretized log-normal (with location parameter 0 and scale parameter $0.7$) and its mirror image. We consider the 10th, 25th, 40th, 50th, 60th, 75th and 90th percentiles of these distributions, all of which lie in the interval $[0,3]$, and re-normalize the corresponding probabilities. The resulting probability distributions are provided in Table \ref{tab:values} below.

\begin{table}[ht]
    \caption{``True'' value distributions}
    \vskip 0.1in
    \label{tab:values}
    \centering
    \begin{tabular}{|c|c|c|c|}
        \hline
         $F(\tau)$ & Uniform & Right-Skewed & Left-Skewed \\ \hline
         $10/ 90$ & 0.250 & 0.194 & 0.278 \\
         $25/ 90$ & 0.625 & 0.357 & 1.426 \\
         $40/ 90$ & 1.000 & 0.543 & 1.900 \\
         $50/ 90$ & 1.250 & 0.700 & 2.100 \\
         $60/ 90$ & 1.500 & 0.902 & 2.257 \\
         $75/ 90$ & 1.875 & 1.374 & 2.443 \\
         $90/ 90$ & 2.250 & 2.522 & 2.606 \\ \hline
    \end{tabular}
\end{table}

Furthermore, we considered three possible ``true'' pricing rules: first-price, second-price, and second-price with a soft-floor equal to $1.0$. We emphasize that these choices are arbitrary and only meant to be illustrative.

To implement our inference procedure, we run 100 iterations (stopping early if search has converged) and report the results for the iteration in which the difference between observed and predicted bids was smallest. Also, in each iteration, we average predicted bids over 10 instances.


\begin{table}[ht]
\caption{MAE of Predicted Percentiles, Uniform}
\label{tab:MAE-uniform}
\vskip 0.1in
\centering
\begin{tabular}{|l|r|r|r|}
\hline
{} &  FirstPrice &  SecondPrice &  softRP\_1.0 \\ \hline
FirstPrice  &    \textbf{0.078647 }&     0.253015 &    0.235419 \\
SecondPrice &    0.260861 &     0.005953 &    \textbf{0.067475} \\
softRP\_1.0  &    0.206956 &     0.028616 &    \textbf{0.012303} \\ \hline
\end{tabular}
\end{table}

\begin{table}[ht]
\caption{MAE of Predicted Percentiles, Right Skewed}
\label{tab:MAE-right}
\vskip 0.1in
\centering
\begin{tabular}{|l|r|r|r|}
\hline
{} &  FirstPrice &  SecondPrice &  softRP\_1.0 \\ \hline
FirstPrice  &    \textbf{0.062768} &     0.502484 &    0.474484 \\
SecondPrice &    0.319719 &     \textbf{0.006436} &    0.110450 \\
softRP\_1.0  &    0.223529 &     0.097096 &    \textbf{0.039177} \\ \hline
\end{tabular}
\end{table}

\begin{table}[H]
\caption{MAE of Predicted Percentiles, Left Skewed}
\label{tab:MAE-left}
\vskip 0.1in
\centering
\begin{tabular}{|l|r|r|r|}
\hline
{} &  FirstPrice &  SecondPrice &  softRP\_1.0 \\ \hline
FirstPrice  &  \textbf{0.058486} &   0.137498 &    0.177831 \\
SecondPrice &  0.215690 & \textbf{0.009505} &  0.102813 \\
softRP\_1.0  &  0.173198 &   0.029954  &    \textbf{0.017729} \\ \hline
\end{tabular}
\end{table}

Tables \ref{tab:MAE-uniform}, \ref{tab:MAE-right} and \ref{tab:MAE-left} report the Mean Absolute Error (MAE)\footnote{We compute this as the arithmetic average of the absolute difference between predicted and actual value for each quantile.} for different choices of ``true'' value distributions and pricing rules. Each table corresponds to a different ``true'' value distribution. Within each table, each row corresponds to a ``true'' pricing rule, and each column to the pricing rule used in the inference procedure---henceforth, the ``hypothesized'' pricing rule.

The main take-away is that the prediction error is  smallest when the hypothesized pricing rule is  the true one; it is also numerically small (see the diagonal in each table). When there is a pricing mismatch, however, the error is larger. This is a useful consistency check for our learning approach. 

\begin{figure}[h]
    \centering
    \includegraphics[scale=0.6]{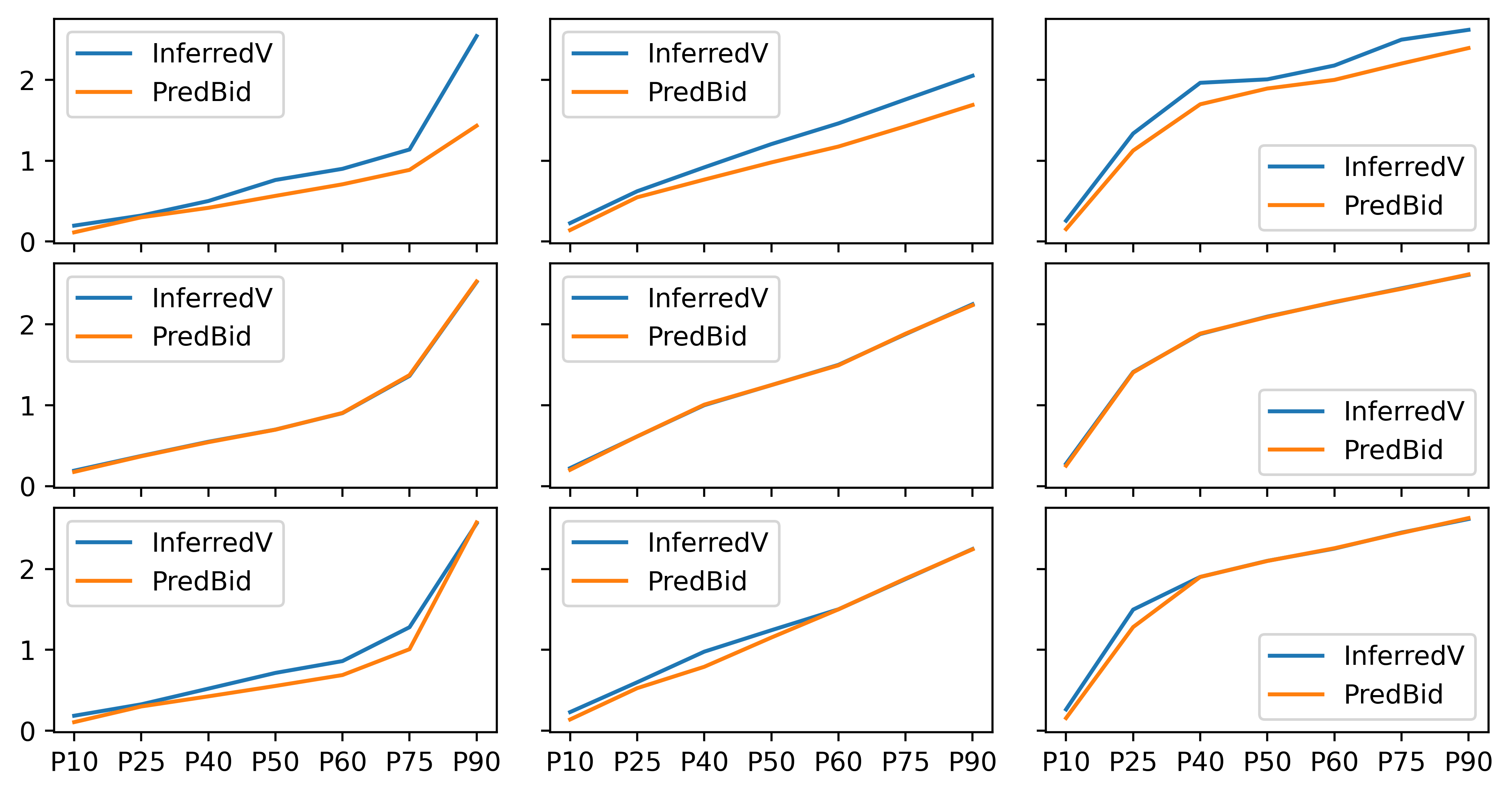}
    \caption{Bid shading effect for V1, V2, V3. Each row corresponds to a different pricing rule, from top to bottom, first-price, second-price, and a \$1.0 soft-floor. Each column corresponds to a different true value distribution, from left to right, left-skewed, uniform, and right-skewed.
    }
    \label{fig:bidshading}
\end{figure}

We also conducted an investigation of \emph{bid shading}, assuming that the hypothesized pricing rule is the actual one. Figure \ref{fig:bidshading} 
summarize
the results. Each line of sub-figures in Figure \ref{fig:bidshading} corresponds to a different pricing rule---from top to bottom, first-price, second-price, and a \$1.0 soft-floor. Each column corresponds to a different true value distribution---from left to right, left-skewed, uniform, and right-skewed. The amount of bid shading was computed based on the \emph{inferred} values and \emph{predicted} bids.

The results for first- and second-price auctions are in line with theoretical expectations. Since this is a single-object auction, there is no evidence of bid shading in second-price auctions, but bids in first-price auctions are largely below the corresponding (inferred) values. The key point is that these theoretical predictions are confirmed even though we use inferred rather than true values (the latter being un-observable in reality). 

Furthermore, we find evidence of bid shading in the presence of a soft-floor. This is especially the case for lower values (left tail), which is in accordance with the fact that the soft-floor turns low-\emph{bid} auctions into first-price auctions, and low-\emph{value} bidders are the ones most likely to place low bids. These results demonstrate that our simulation approach is able to infer bid shading even in the presence of complex or ``non-textbook'' pricing rules.

We note that \citet{nekipelov2015econometrics} finds evidence of bid shading in Bing search ad data. The authors analyze the data assuming it comes from a generalized second-price auction. However, current Microsoft documentation does not explicitly indicate the pricing rule.\footnote{See \url{https://help.ads.microsoft.com/\#apex/ads/en/53099/0}.} By way of contrast, our simulation analysis generates ``observed bids''  by using the second-price rule. Correspondingly, we find no bid shading in second-price auctions even using inferred values.


\subsubsection{Inferring values in  production environment}
\label{prod}
Finally, we use our approach to infer the distribution of bidders' values from the observed distribution of bids in a production environment. 
The analysis is based on aggregated bid data for two specific shopper queries in an e-commerce setting, one characterized by low traffic and the other by high traffic. The data aggregation process converts all bids into a bid per impression, so we set all click-through rates to 1. Thus, we apply our analysis to a symmetric environment with a single query, unit click-through rates, and two different scenarios, one with a low number of bidders (low traffic) and one with a high number of bidders (high traffic). 

We normalize all bids to lie in a grid $B$ in the interval $[0,4]$. Specifically, we take $B = \{\frac{i}{10} : i = 0,\ldots,40\}$. We also define the set of types to be quantiles of the given bid distribution, with a step size of $0.1$.
As in the previous  subsection, we initialize our iterative procedure by setting the \emph{inferred} \emph{value} distribution to be the \emph{observed} \underline{bid} distribution, except that now the latter are real data from the noted e-commerce setting. We then apply multiple iterations of the procedure described in \eqref{eq:inference}, with a learning rate adjustment of $\alpha=0.2$. In every inference iteration, since the inferred values are associated with increasing quantiles, we apply a ``flattening'' step to ensure that they are indeed increasing.

\begin{figure}[ht]
    \centering
    \includegraphics[width=\textwidth]{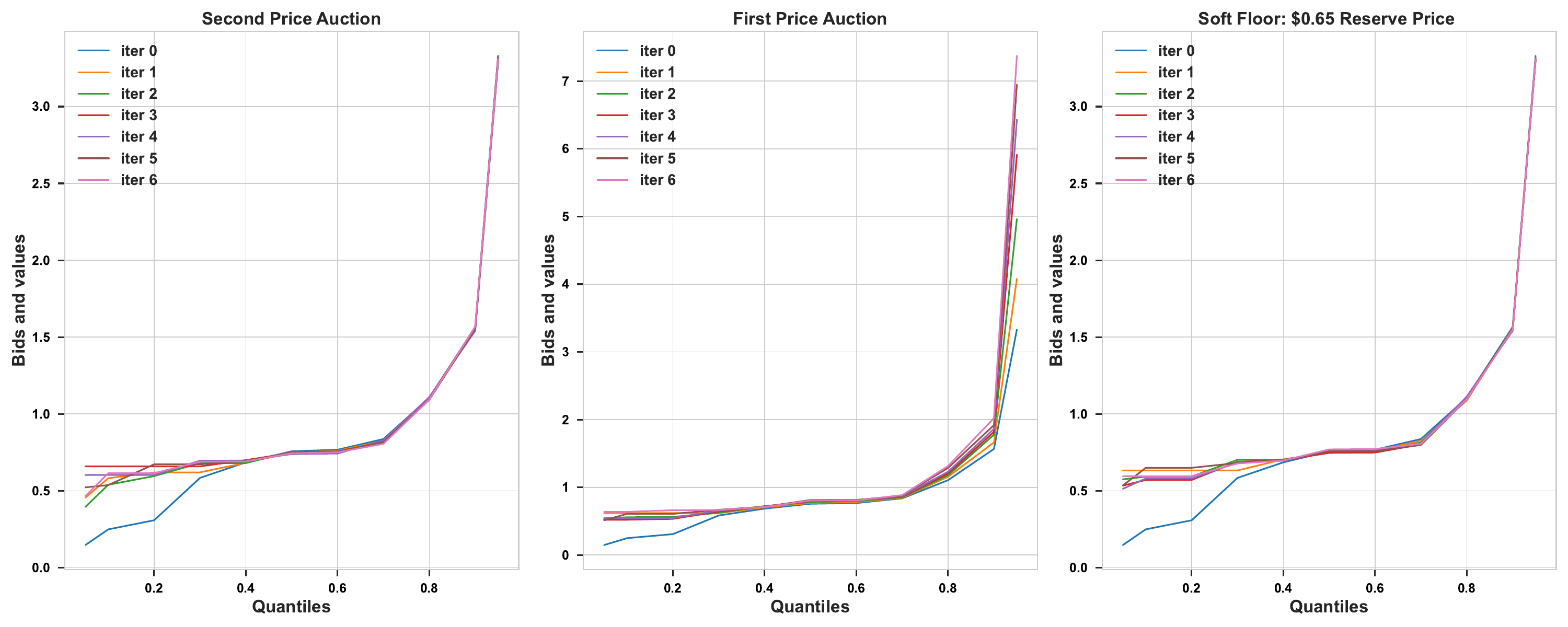}
    \caption{Value inference in three auction formats for the low traffic search query}
    \label{fig:record-sleeve-prod}
\end{figure}

Figure \eqref{fig:record-sleeve-prod}  represents the inferred values of bidders for 
the low-traffic search query,
under the three pricing rules we analyzed in \S \ref{sec:results:SFRP}: first price, second price, and soft-floor with a reserve price of $\$0.65$. We chose these pricing rules arbitrarily, to demonstrate their impact on the inference process. We used 8 iterations of the inference procedure and 3 runs of the learning algorithm per iteration, each with $T = 500,000$, averaging bids over the last $50,000$ periods.

\begin{figure}[htb]
\centering
    \begin{minipage}{.5\textwidth}
        \centering
        \includegraphics[width=1\textwidth]{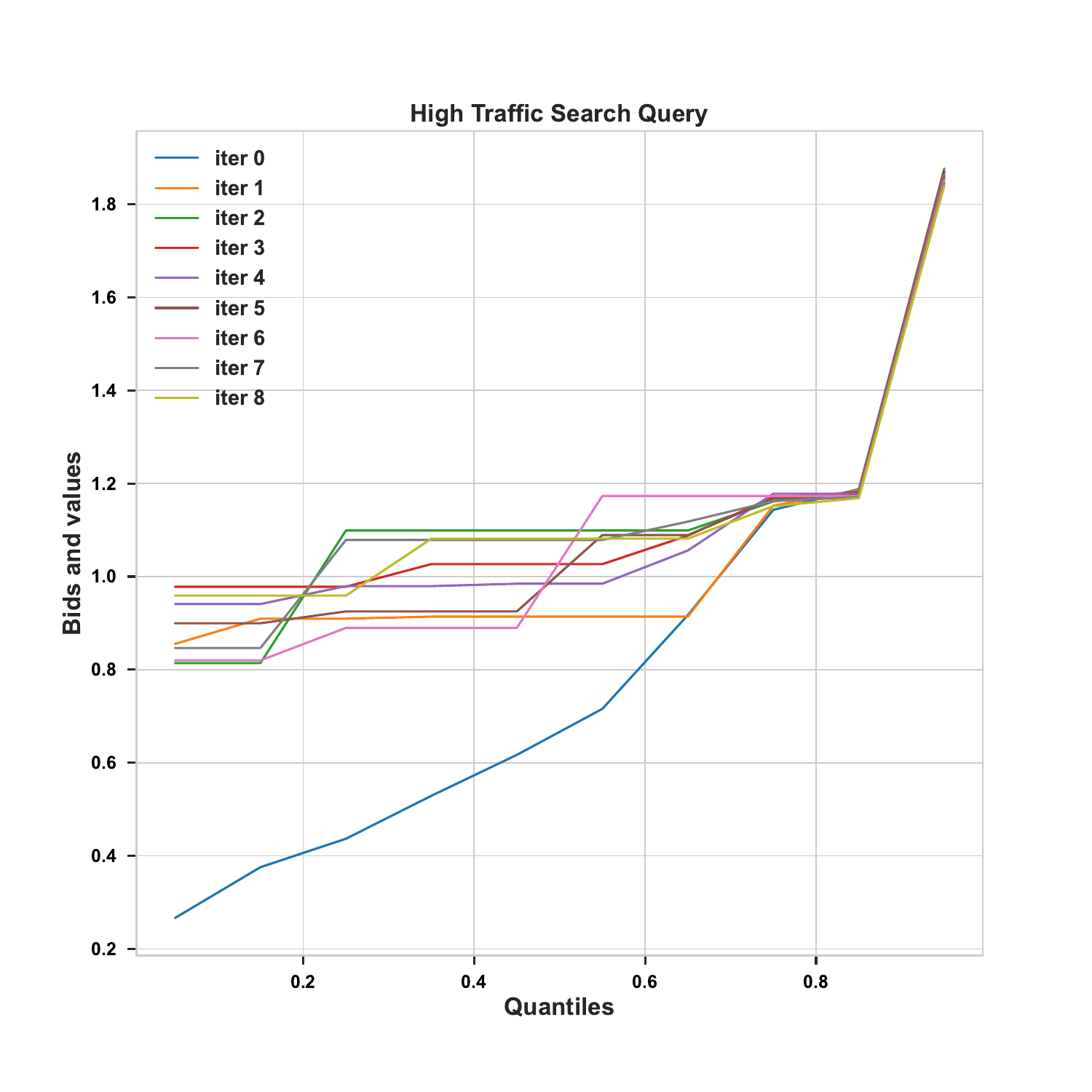}
    \end{minipage}
    \begin{minipage}{.4\textwidth}
    \caption{Inferring values for the high traffic search query. In this experiment, we set $T=800,000$ and applied $8$ inference iterations. The results show that inferred values converge in a few iterations even in a more realistic auction environment.}
    \label{fig:bid-shading-prod}
    \end{minipage}
\end{figure}

As anticipated, we observe bid shading in the first price auction, as well as within the lower to middle quantiles in the case of the soft-floor. The second price auction also displays bid shading at lower quantiles. We hypothesize that this deviation from theoretical prediction is due to a lack of learning amongst low-valuation types: players with low valuation win rarely, so the feedback they receive is coarse on most periods and hence insufficient to converge to bidding one's value.\footnote{We observe the same pattern of deviations from truthful bidding for low types in the second-price auction studied in \S \ref{sec:results:textbook}.} Figure \ref{fig:bid-shading-prod} analyzes the 
high-traffic query. We increased the length of the simulation to $T=800,000$ periods. As observed, even with a large number of bidders,
inferred values converge in a few iterations.\footnote{We used a realistic pricing function, but for confidentiality reasons we are unable to provide details.}


\raggedbottom
\section{Conclusions}
\label{sec:conclusions}

We demonstrated that online learning algorithms can serve as effective tools in modeling bidders in complex ad auctions where  it is not feasible to solve for equilibrium behavior. We showed that soft floors can have a positive effect on the ad service's revenue, even in an ex-ante symmetric environment, provided values and click-through rates are query-dependent, but yield lower revenue than aptly chosen standard reserve prices in single-query environments with a dominant bidder. We also documented that the algorithm choice can significantly affect the learning rate in a more realistic auction environments, especially with a large number of bidders.
Utilizing our simulation approach, we also showed how to infer bidders' valuations in the presence of more realistic auction rules. We demonstrated this using aggregate bid data from an e-commerce website in both low- and high-density auctions.

\raggedbottom
\section{Acknowledgments}
\label{sec:ack}

The authors would like to thank Ben Allison, Pat Copeland, Olivier Jeunen, Lihong Li, Muthu Muthukrishnan, Amin Sayedi, Pooja Seth, Yongning Wu, Eva Yang, Ruslana Zbagerska, Zuohua Zhang,
and the reviewers for their support and insightful feedback.

\bibliography{refs}
\bibliographystyle{abbrvnat}

\end{document}